%% file: main.tex
\documentclass[10pt,twocolumn,letterpaper]{article}

\usepackage[pagenumbers]{cvpr} 

\usepackage[dvipsnames*,svgnames]{xcolor}
\usepackage{colortbl}
\usepackage{multirow}
\usepackage{adjustbox}
\usepackage{caption}
\usepackage{MnSymbol}

\definecolor{cvprblue}{rgb}{0.21,0.49,0.74}
\usepackage[pagebackref,breaklinks,colorlinks,citecolor=cvprblue]{hyperref}
\usepackage{multicol}
\usepackage{algorithm}
\usepackage{algorithmic}
\usepackage{fontawesome}


\definecolor{cvprblue}{rgb}{0.21,0.49,0.74}
\usepackage[pagebackref,breaklinks,colorlinks,citecolor=cvprblue]{hyperref}

\def\paperID{12} 
\def\confName{CVPR}
\def\confYear{2024}

\title{SciFlow: Empowering Lightweight Optical Flow Models with \\Self-Cleaning Iterations} 

\author{Jamie Menjay Lin$^1$~~~
Jisoo Jeong$^2$~~~
Hong Cai$^2$~~~
Risheek Garrepalli$^2$~~~ 
Kai Wang$^1$~~~ 
Fatih Porikli$^2$\\
$^1$Qualcomm Technologies, Inc. ~~~
$^2$Qualcomm AI Research$^{\dagger}$\\
{\tt\small \{jmlin, jisojeon, hongcai, rgarrepa, kwang, fporikli\}@qti.qualcomm.com}
}

\begin{document}
\maketitle
\input{sec/0_abstract}    
\input{sec/1_intro}
\input{sec/2_related}

\input{sec/3_method}
\input{sec/4_experiments}


\input{sec/6_conclusion}

\newpage

{
    \small
    \bibliographystyle{ieeenat_fullname}
    \bibliography{main}
}


\end{document}

%% file: sec/0_abstract.tex
\begin{abstract}
Optical flow estimation is crucial to a variety of vision tasks. Despite substantial recent advancements, achieving real-time on-device optical flow estimation remains a complex challenge. First, an optical flow model must be sufficiently lightweight to meet computation and memory constraints to ensure real-time performance on devices. Second, the necessity for real-time on-device operation imposes constraints that weaken the model's capacity to adequately handle ambiguities in flow estimation, thereby intensifying the difficulty of preserving flow accuracy.

This paper introduces two synergistic techniques, {Self-Cleaning Iteration (SCI)} and {Regression Focal Loss (RFL)}, designed to enhance the capabilities of optical flow models, with a focus on addressing optical flow regression ambiguities. These techniques prove particularly effective in mitigating error propagation, a prevalent issue in optical flow models that employ iterative refinement. Notably, these techniques add negligible to zero overhead in model parameters and inference latency, thereby preserving real-time on-device efficiency. 

The effectiveness of our proposed SCI and RFL techniques, collectively referred to as SciFlow for brevity, is demonstrated across two distinct lightweight optical flow model architectures in our experiments. Remarkably, SciFlow enables substantial reduction in error metrics (EPE and Fl-all) over the baseline models by up to 6.3\% and 10.5\% for in-domain scenarios and by up to 6.2\% and 13.5\% for cross-domain scenarios on the Sintel and KITTI 2015 datasets, respectively.
\end{abstract}

%% file: sec/1_intro.tex
\section{Introduction}
\label{sec:intro}

\begin{figure}[t]
\begin{center}$
\centering
\begin{tabular}{c c c c }
 & \text{\small{(a) Base Model}} & \hspace{-0.35cm} \text{\small{(b) Base+SCI}} & \hspace{-0.35cm} \text{\small{(c) Base+SCI+RFL}}\\

\vspace{-0.1cm} \hspace{-0.2cm}  \rotatebox{90}{\ \quad  \text{\textbf{\textit{iter = 1}}}}
& \hspace{-0.35cm} \includegraphics[width=2.5cm]{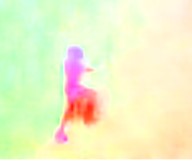} 
& \hspace{-0.4cm} \includegraphics[width=2.5cm]{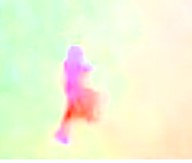} 
& \hspace{-0.4cm} \includegraphics[width=2.5cm]{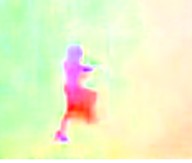} \\
\vspace{-0.1cm} \hspace{-0.2cm} \rotatebox{90}{\ \quad \text{\textbf{\textit{iter = 3}}}}
& \hspace{-0.35cm} \includegraphics[width=2.5cm]{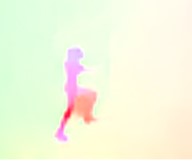} 
& \hspace{-0.4cm} \includegraphics[width=2.5cm]{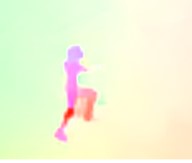}
& \hspace{-0.4cm} \includegraphics[width=2.5cm]{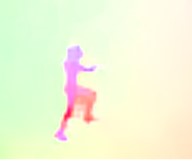}\\
\vspace{-0.1cm} \hspace{-0.2cm} \rotatebox{90}{\ \quad \text{\textbf{\textit{iter = 5}}}}
& \hspace{-0.35cm} \includegraphics[width=2.5cm]{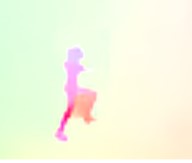} 
& \hspace{-0.4cm} \includegraphics[width=2.5cm]{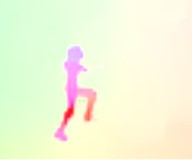} 
& \hspace{-0.4cm} \includegraphics[width=2.5cm]{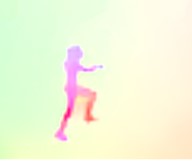}\\
\vspace{-0.1cm} \hspace{-0.2cm} \rotatebox{90}{\ \quad \text{\textbf{\textit{iter = 7}}}}
& \hspace{-0.35cm} \includegraphics[width=2.5cm]{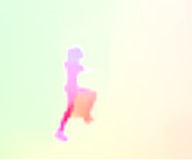} 
& \hspace{-0.4cm} \includegraphics[width=2.5cm]{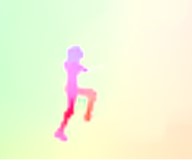} 
& \hspace{-0.4cm} \includegraphics[width=2.5cm]{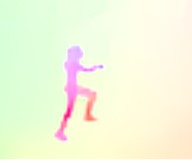}
\end{tabular}$
\end{center}
\vspace{-10pt}
\caption{\textbf{A zoomed-in demonstration of "Self-Cleaning Iterations (SCI)" effect against error propagation, a prevalent issue in iterative refinement for optical flow models.} (a) The baseline model (RAFT-Small \cite{teed2020raft} as one choice of model architecture) suffers from error propagation over iterations, especially near the arm and legs. (b) When the SCI technique is applied to the baseline model, it demonstrates a ``self cleaning" effect over iterations. This is achieved at negligible additional overhead in computation and in model size. (c) When both the SCI and RFL techniques are applied to the baseline model, the ``self cleaning" effect becomes even more visible, particularly around the arm and feet. On top of "Base+SCI", this RFL technique concerns only the loss function in training so it adds no additional overhead for inference.}
\label{exp:intro}
\vspace{-10pt}
\end{figure}

{\let\thefootnote\relax\footnotetext{{
\hspace{-6.5mm} $\dagger$ Qualcomm AI Research is an initiative of Qualcomm Technologies, Inc.}}}

Optical flow is a fundamental task that represents pixel-level correspondence between two consecutive video frames. Since optical flow provides pixel-level movement, it is widely used for a variety of video perception tasks, e.g., action recognition~\cite{lee2018motion, cai2019temporal}, object tracking~\cite{kale2015moving, zhou2018deeptam}, video compression~\cite{wu2018video, lu2019dvc}, video frame interpolation~\cite{kong2022ifrnet, jeong2024ocai}. 

Thanks to the recent advances in deep learning, optical flow estimation models have become significantly more accurate by leveraging neural networks~\cite{ilg2017flownet, teed2020raft, huang2022flowformer}. While some earlier works lack a principled way to design neural networks for capturing pixel correspondences, more recently, RAFT~\cite{teed2020raft} have proposed an optimization-inspired architecture that sets a new baseline for the method and the model architecture. More specifically, it constructs a global and re-usable cost volume with pairwise correlations between extracted image features of the two frames and then uses a recurrent neural module to iteratively refine the optical flow. This essentially mimics the optimization steps used conventional computer vision algorithms for solving correspondences, and has been shown to greatly improve accuracy and generalizability. As a result, most subsequent and current state-of-the-art solutions follow a similar model design strategy.

During the iterative estimation process, the predicted optical flow is prone to errors especially in the earlier stages or under high ambiguity. While the iterations can rectify many errors, other errors in some cases could persist and even be further propagated into later iterations, impacting final model accuracy. Figure~\ref{exp:intro} provides an example of this. In the top row for the first iteration, the initial flow estimates tend to be not accurate, with apparent errors near the person's arm and legs as well as in the background. Through the iterations, the estimation on the background pixels improves, but other errors near the legs persist and the estimation on the arm becomes even less accurate. There is generally a lack of an effective solution to handle such issue of error propagation for many optical flow methods.


In this paper, we propose a novel and effective approach, Self-Cleaning Iterations (SCI), to address the issue of error propagation that is often observed during the iterative refinement process of optical flow models. 
We enable the network to ``self-assess" the likely correctness of flow estimates during the iterative refinement.
More specifically, in each iteration, we compare the feature maps of the two frames using the current estimated optical flow and warping. The pixel-wise differences provide an indication for consistency of the optical flow, which are converted into a quality range between 0 and 1. The resulting dense quality measure is consumed by the model as an additional feature channel to guide the network to ``self-correct" inconsistencies in next iterations.

In addition, during training, we introduce a new loss, namely, Regression Focal Loss (RFL), to better leverage the available ground truth to improve the network's awareness for regions of potentially incorrect estimates. Existing optical flow training schemes predominantly weight the pixel-wise loss equally, without taking into account the different prediction accuracy on each pixel at a given iteration. In contrast, our RFL gives heavier weights to regions of high residual regression errors, encouraging the network to focus its learning more on regions where it faces higher ambiguities to find feature correspondences.

Our proposed techniques add negligible or zero computation overhead at inference time, which is particularly critical for lightweight optical flow models intended for real-time on-device targets, such as mobile phones and AR/VR devices. Specifically, SCI only requires the network to process an additional channel of the quality map and RFL only affects loss computation during training. In contrast, many of the latest state-of-the-art methods require more complex computations for accuracy improvement, including heavier models or transformer architecture. Despite being parsimonious on computation usage, our proposed approach effectively improves multiple baseline architectures, achieving the best accuracy when comparing to other existing lightweight optical flow models.

Our main contributions are summarized as follows:
\begin{itemize}

\item We propose a novel technique, Self-Cleaning Iteration (SCI), which enables the model to ``self-assess" flow quality in current iteration, and to ``self-clean" flow estimates in subsequent iterations for optical flow models. This helps resolve ambiguities in the estimation and mitigates error propagation during the iterative refinement. It is noteworthy that SCI incurs minimal computational overhead during model inference.

\item In addition, we propose a Regression Focal Loss (RFL), which guides the model to focus more on regions of high residual regression errors, thus encouraging the model to learn better to improve for those challenging scenarios where feature correspondences are harder.

\item We further combine both techniques, SCI and RFL and verify our proposal on two distinct optical flow baseline architectures. Our experiments demonstrate that our SCI and RFL jointly serve as an effective unified solution to handle ambiguities in both in-domain and cross-domain scenarios. Remarkably, our solution also leads to state-of-the-art accuracy results compared with existing lightweight optical flow models.


\end{itemize}

\begin{figure*}[!t]
\centering
\includegraphics[width=0.99\linewidth]{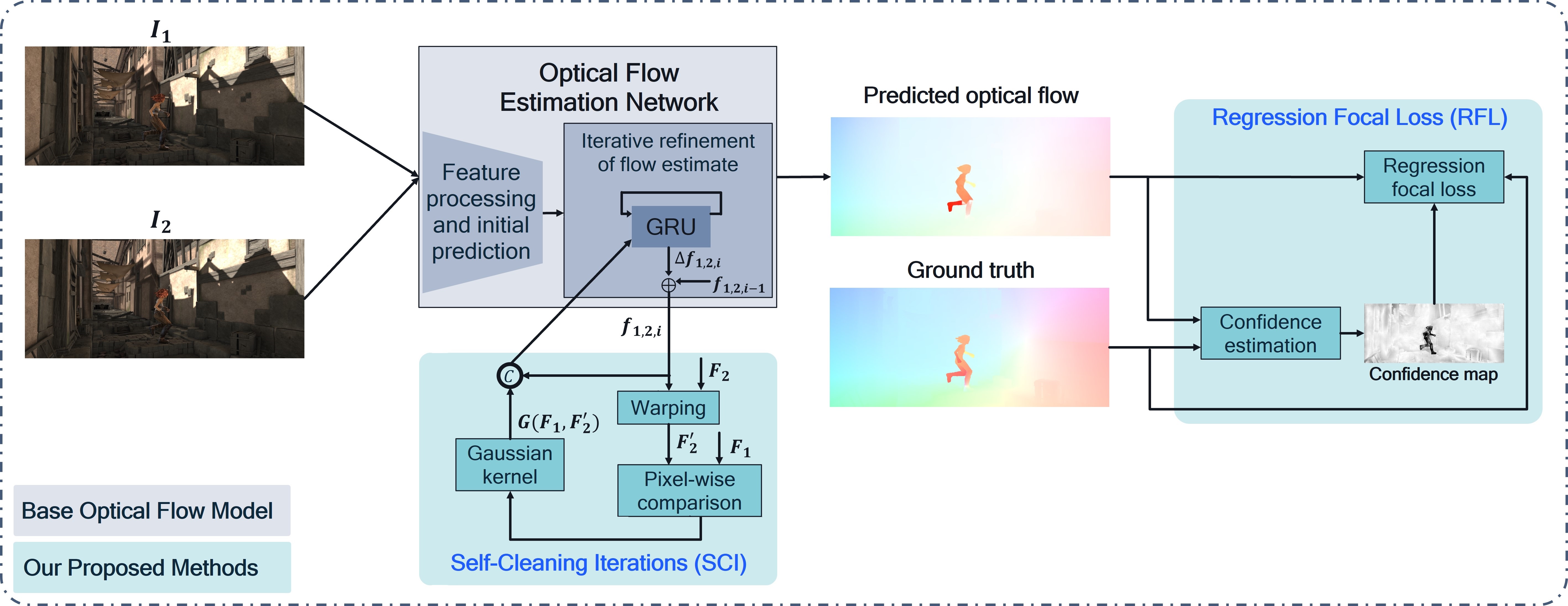}
\vspace{-5pt}
\caption{\textbf{Overview of our proposed approach.} \textbf{Self-Cleaning Iterations (SCI)} enables the network to ``self-assess" the flow prediction quality and then to ``self-clean" the flow prediction itself over the standard practice of iterative refinement process in many optical flow models. \textbf{Regression Focal Loss (RFL)} derives a confidence map and guide the network to focus more on regions of high residual regression errors during the iterations.
${\small{\textcircled{\scriptsize{C}}}}$ stands for the concatenation operator.
}
\vspace{-0pt}
\label{fig:overview}
\vspace{-3mm}
\end{figure*}


%% file: sec/2_related.tex
\section{Related Work}
\label{sec:related}

\subsection{Optical Flow Models with Iterative Refinement}




RAFT~\cite{teed2020raft} introduced a new optical flow model design and has since become the baseline architecture for many later advancements. It builds a cross-level global correlation volumes and iteratively refines the prediction using convolutional gated recurrent units (ConvGRU) \cite{cho2014learning}. GMA \cite{jiang2021learning}, Flowformer \cite{huang2022flowformer}, and FlowFormer++ \cite{shi2023flowformer++} among others keep improving model accuracy while keeping this ConvGRU baseline design.

Despite the success of ConvGRU being an effective neural ODE optimizer \cite{ChenRBD18, NEURIPS2019_455cb265, teed2020raft}, the error propagation behavior due in part to irresolvable ambiguities in estimates has not been explicitly discussed. In this paper, we aim at mitigating this issue prevalent in ConvGRU-based models.

\subsection{Uncertainty-Aware Optical Flow Estimation}

Several prior works have explored incorporating confidence or uncertainty estimates into their models \cite{gast2018lightweight,truong2021learning,yin2019hierarchical, jeong2023distractflow} among others. For instance, \cite{gast2018lightweight} modify the network's output layer to predict variance at intermediate layers and use assumed density filtering to propagate uncertainty across the network. \cite{truong2021learning} (PDC-Net) take a probabilistic approach, employing a mixture distribution for prediction and a separate uncertainty decoder within their multi-stage architecture to decouple flow estimation from uncertainty estimation. While PDC-Net is the closest work to ours, we differ in two key aspects. First, we do not explicitly use a dedicated uncertainty decoder. Second, we leverage geometric consistency for confidence estimation and utilize a self-cleaning mechanism for our iterative refinement.

\subsection{Standard Loss Function for Optical Flow}

FlowNet~\cite{dosovitskiy2015flownet} uses end-point error loss, mathematically the Euclidean distance, between the ground truth and predicted flow. PWC-Net~\cite{sun2018pwc} uses L1 and L2 losses. L2 loss is first applied in the initial stage of training, while L1 loss is applied in subsequent finetuning. Several recent optical flow models~\cite{teed2020raft, huang2022flowformer} apply iterative refinement by summing weighted L1 losses over multiple iterations. 
\begin{equation} \label{eq:optical_each_l1}
l_\text{i} = ||f_\text{gt} - f_\text{i}||_1
\end{equation}
where $f_{gt}$ and $f_{i}$ are the optical flow ground truth and prediction, respectively, in iteration $i$. A scheme below for weighted combination is then used over multiple iterations. 
\begin{equation} \label{eq:optical_l1}
\mathcal{L}_\text{total} = \sum^{N}_{i=1} \gamma ^{N-i} \cdot l_\text{i}
\end{equation}
where N stands for the iteration index and $0<\gamma<1$ is a decay factor over iterations. The whole predicted flow map of each iteration is weighted accordingly before accumulation.

Several other works adopt complementary insights or regularization objectives based on semantic segmentation, object depths, multi-frame aggregation, temporal consistency, occlusion consistency, or transformation consistency \cite{DBLP:conf/cvpr/JeongLPK22, cai2021x, Borse_2023_CVPR, yasarla2024futuredepth, yasarla2023mamo, das2023transadapt} to further enhance their model accuracy on top of the standard optical flow loss function.

\subsection{Focal Loss Function for Dense Classification}

Focal Loss~\cite{lin2017focal} has been proven an effective technique to address the class imbalance issue in dense classification tasks, such as segmentation.
It places higher emphasis on feature samples of less (or under) represented classes. 
\begin{equation} \label{eq:crossep_loss}
CE(p_t) = -log(p_t)
\end{equation}
\begin{equation} \label{eq:focal_loss}
FL(p_t) = -(1-p_t)^\gamma log(p_t)
\end{equation}
where CE and FL represent the cross entropy and focal loss, respectively. And, $p_t$ is the probability for a class.
Focal loss is originally proposed for dense classification tasks and is designed to work with the cross entropy loss. The original form of focal loss is not directly applicable to optical flow estimation, a regression task without the definition of classes. Moreover, cross entropy is not used in the standard loss for optical flow estimation.


\subsection{Lightweight Optical Flow Models}


\cite{garrepalli2023dift,zhang2024neuflow,kong2021fastflownet} among others are recent works on lightweight design with attention or cost volume construction in a coarse-to-fine paradigm. \cite{garrepalli2023dift} is a lightweight version of \cite{teed2020raft}, which adopts single level cost volume per iteration and adopts coarse-to-fine cost volumes with finest resolution being $1/16$ and demonstrates real time performance on Snapdragon\textsuperscript{\faRegistered} 8 Gen 1 HTP. \cite{zhang2024neuflow} 
first performs global matching at $1/16$ resolution and then refines flow at $1/8$ using lightweight CNN layers and demonstrates real time performance on Jetson Orion Nano. \cite{kong2021fastflownet} 
also adopts coarse-to-fine and in addition uses dilated correlation layer for lighter cost volume. In addition there were additional previous works addressing light weight optical flow estimation like \cite{hui2020liteflownet3, ranjan2017optical}.

\vspace{-2mm}

\begin{figure}[!t]
\centering
\includegraphics[width=0.80\linewidth]{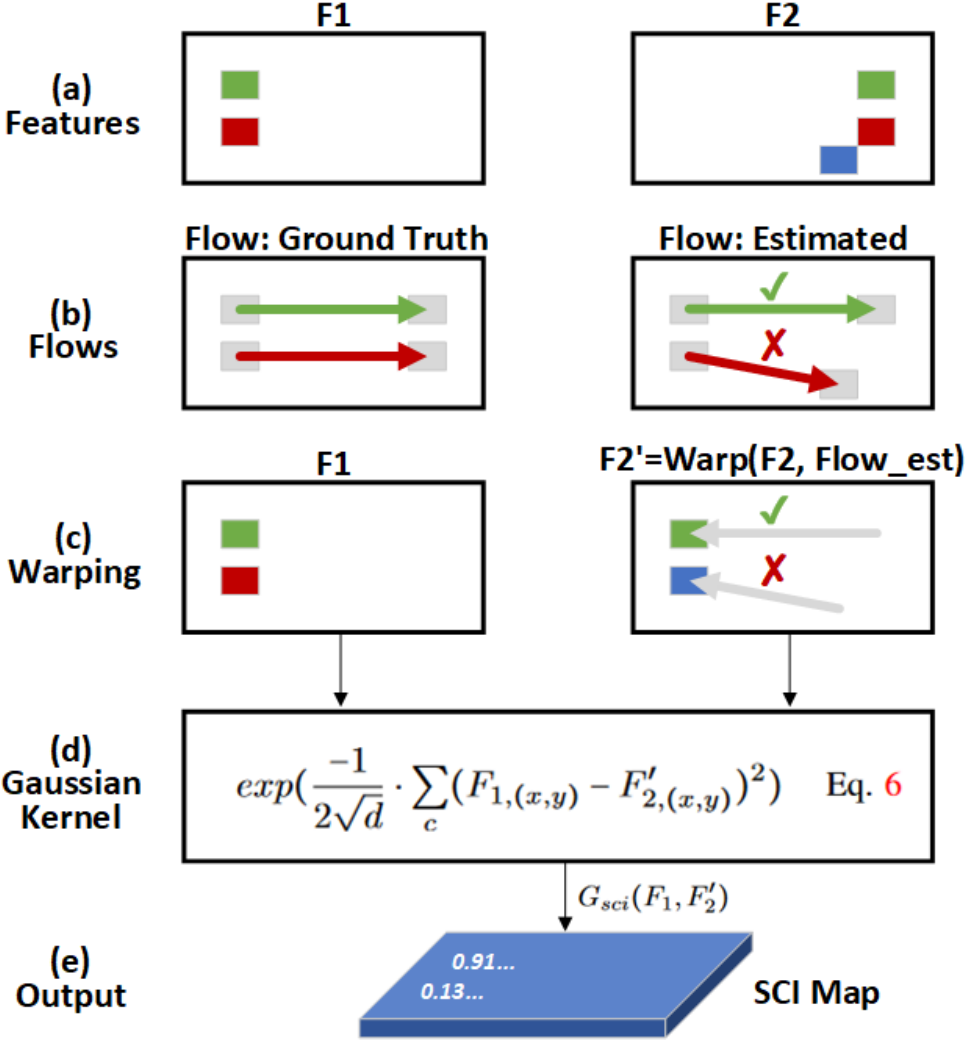}
\vspace{-5pt}
\caption{\textbf{Concept for SCI map creation.} (a) A pair of image features are taken as input. (b) The ground truth flows point to their matches on the left sub-figure, while the estimated flows point incorrectly for some features on the right sub-figure. (c) F2' is derived by warping F2 by the estimated flows. (d) F1 and F2' are taken by their tensor-wise differences for Gaussian Kernel (Eq. \ref{eq:gaussian}) evaluation for their affinity. (e) A dense SCI map output is derived.}
\vspace{-5pt}
\label{fig:sci}
\end{figure}


%% file: sec/3_method.tex
\section{Method}
\label{sec:method}
\vspace{-1mm}

In this section, we discuss details of these two interrelated methods, Self-Cleaning Iteration (SCI) and Regression Focal Loss (RFL). The first technique, SCI, is applied for both training and inference by actively computing the similarity between the reference frame and the warped frame based on the flow estimate. RFL, the second technique as a loss function in a similar arithmetic formulation to that of SCI, is proposed to guide model learning by focusing more on regions of high residual regression errors. Figure \ref{fig:overview} gives an overview for the system setup, including an optical flow estimation network and our proposed SCI and RFL methods.

\subsection{Self-Cleaning Iterations}

In this section, we present the concept of Self-Cleaning Interactions (SCI) in the first half, and then the details of the method in the second half. Based on our observations in many iterative refinement-based models, we notice that errors made in early iterations of estimation could persist through subsequent iterations, affecting the quality of the dense flow estimates. To address this issue, SCI is designed to assess the quality of these estimates.

\begin{figure}[!t]
\centering
\includegraphics[width=1.0\linewidth]{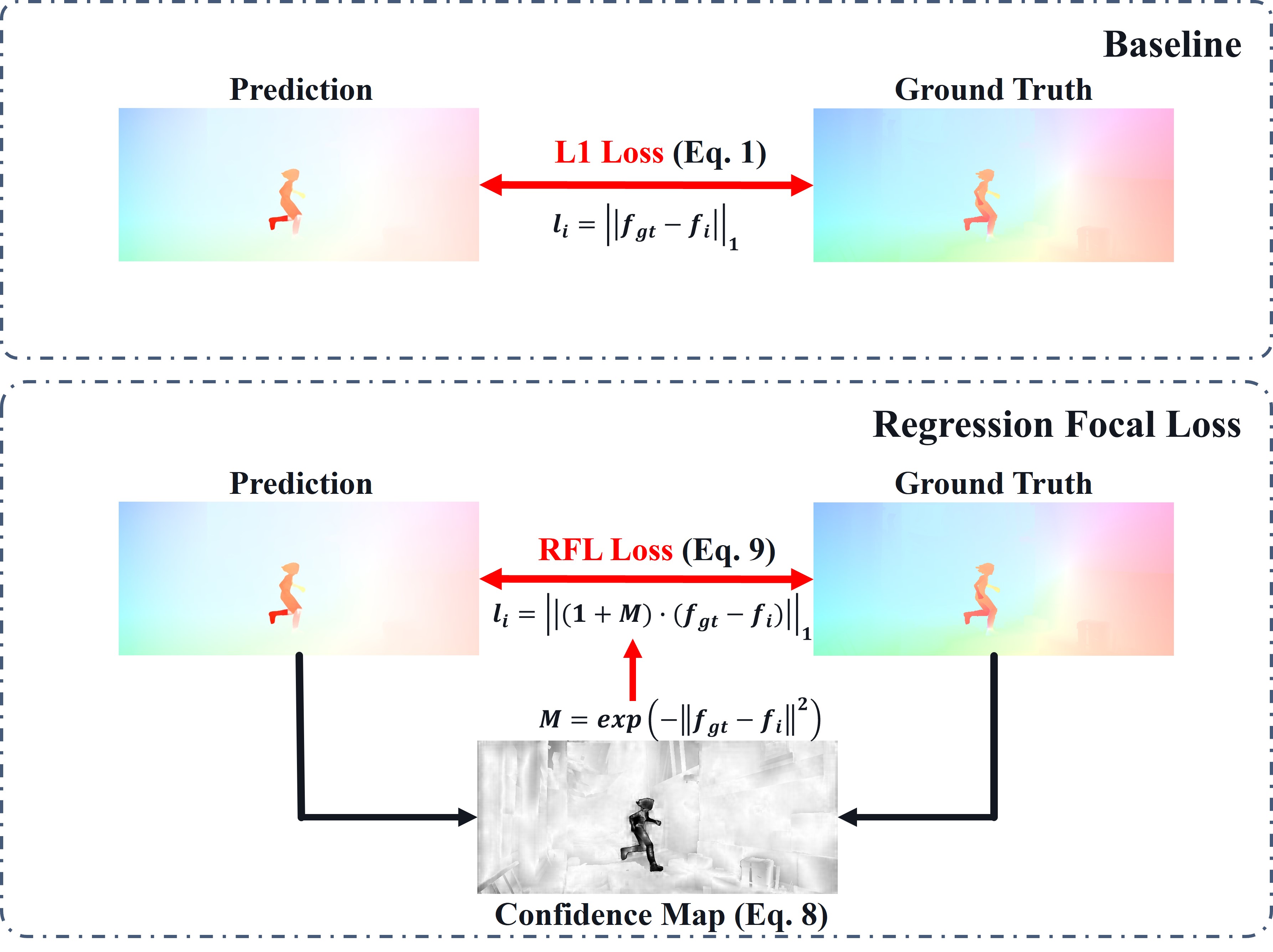}\\
\vspace{-10pt}
\caption{\textbf{Regression Focal Loss.}
While equal-weight loss across all pixels is used for conventional optical flow model training (Eq.~\ref{eq:optical_each_l1}), Regression Focal Loss generates the confidence map using optical flow prediction and ground truth (Eq.~\ref{eq:conf}) and leverages it to the optical flow loss (Eq.~\ref{eq:rfl_l1}) so that model can focus on difficult areas in the dataset.
}
\vspace{-5pt}
\label{fig:rfl}
\end{figure}
\begin{table*}[t]
\begin{center}
\caption{\textbf{Cross-domain and in-domain optical flow estimation results on Sintel (train) and KITTI (train) datasets.} All models of cross-domain in this table are trained on FlyingChairs (C) and FlyingThings (T), following train protocol in \cite{teed2020raft}. All models of in-domain in this table are finetuned on Sintel (S) and KITTI (K) using pre-trained model (C+T), following \cite{teed2020raft} training protocol.
}
\vspace{-3mm}
\label{tab:main_table}
\adjustbox{max width=1.0\textwidth}
{
\begin{tabular}{|l|c||l|c||c|c|c|c||c|c|c|}
\hline
\multirow{2}{*}{} & Training & \multirow{2}{*}{Method }  & \multirow{2}{*}{$\#$ Params} & \multicolumn{2}{|c|}{Sintel (train)} & \multicolumn{2}{|c|}{KITTI (train)} & \multicolumn{2}{|c|}{Sintel (test)} & KITTI (test) \\
\cline{5-11}
& Datasets &  & & Clean & Final & EPE & Fl-all & Clean & Final & Fl-all \\
\hline
\hline
\multirow{9}{*}{\rotatebox{90}{\textbf{Cross-Domain}}} & \multirow{9}{*}{C+T} 
& PWC-Net \cite{Sun2018PWC-Net} & 8.8 M & 2.55 & 3.93 & 10.35 & 33.70 & - & - & - \\
& & LiteFlowNet2 \cite{hui2020lightweight} & 6.4 M & 2.24 & 3.78 & 8.97 & 25.90 & - & - & - \\
& & LiteFlowNet3 \cite{hui2020liteflownet3} & 5.2 M & 2.59 & 3.91 & 10.40 & - & - & - & - \\
& & FDFlowNet \cite{kong2020fdflownet} & 5.8 M & 2.60 & 4.12 & 10.75 & 29.59 & - & - & - \\
& & FastFlowNet \cite{kong2021fastflownet} & 1.4 M & 2.89 & 4.14 & 12.24 & 33.10 & - & - & - \\
& & MaskFlowNet-small \cite{zhao2020maskflownet} & - & 2.33 & 3.72 & - & 23.58 & - & - & - \\
& & DICL \cite{wang2020displacement} & - & 1.94 & 3.77 & 8.70 & 23.60 & - & - & - \\
& & DIFT \cite{garrepalli2023dift} & - & 3.11 & 4.19 & 12.87 & 43.83 & - & - & - \\
\cline{3-11}
& & $\text{MobileFlow}$\textsuperscript{\ref{mobileflow}} & 1.5 M &  1.79 (-0.0\%) & 3.47 (-0.0\%) & 8.33 (-0.0\%) & 22.06 (-0.0\%) & - & - & - \\
& &\cellcolor{Lavender}$\text{MobileFlow}$\textsuperscript{\ref{mobileflow}}+SCI+RFL (Ours)  & \cellcolor{Lavender}1.5 M & \cellcolor{Lavender}\textbf{1.68 (-6.2\%)} & \cellcolor{Lavender}\textbf{3.34 (-3.8\%)} & \cellcolor{Lavender}\textbf{7.21 (-13.5\%)} & \cellcolor{Lavender}\textbf{20.75 (-5.9\%)} & \cellcolor{Lavender}- & \cellcolor{Lavender}- & \cellcolor{Lavender}- \\
\hline
\hline
 \multirow{8}{*}{\rotatebox{90}{\textbf{In-Domain}}} & \multirow{8}{*}{C + T + S/K}
& PWC-Net \cite{Sun2018PWC-Net} & 8.8 M & 2.02 & 2.08 & 2.16 & 9.80 & 4.39 & 5.04 & 9.60 \\
& & LiteFlowNet2 \cite{hui2020liteflownet3} & 6.4 M & 1.41 & 1.83 & 1.33 & 4.32 & 3.48 & 4.69 & 7.62 \\
& & LiteFlowNet3 \cite{hui2020liteflownet3} & 5.2 M & 1.43 & 1.90 & 1.39 & 4.35 & 2.99 & 4.45 & 7.34 \\
& & FDFlowNet \cite{kong2020fdflownet} & 5.8 M & 1.80 & 1.93 & 1.56 & 6.36 & 3.71 & 5.11 & 9.38 \\
& & FastFlowNet \cite{kong2021fastflownet} & 1.4 M & 2.08 & 2.71 & 2.13 & 8.21 & 4.89 &  6.08 & 11.22 \\
& & DDCNet (B1) \cite{salehi2023ddcnet} & 3.0 M & 1.96 & 2.25 & 2.57 & 15.56 & 6.19 & 6.91 & 38.23 \\
\cline{3-11}
%
\cline{3-11}
& & $\text{MobileFlow}$\textsuperscript{\ref{mobileflow}} & 1.5 M &  1.09 (-0.0\%) & 1.76 (-0.0\%) & 0.96 (-0.0\%) & 3.14 (-0.0\%) & - 
 & - & - \\
& & \cellcolor{Lavender}$\text{MobileFlow}$\textsuperscript{\ref{mobileflow}}+SCI+RFL (Ours)  & \cellcolor{Lavender}1.5 M & \cellcolor{Lavender}\textbf{1.03 (-5.5\%)} & \cellcolor{Lavender}\textbf{1.65 (-6.3\%)} & \cellcolor{Lavender}\textbf{0.92 (-4.2\%)} & \cellcolor{Lavender}\textbf{2.81 (-10.5\%)} & \cellcolor{Lavender}\textbf{2.62} & \cellcolor{Lavender}\textbf{3.80} & \cellcolor{Lavender}\textbf{5.82} \\
 \hline
\end{tabular}
}
\vspace{-8mm}
\vspace{6pt}
\end{center}
\end{table*}

The core intuition behind SCI is the concept of ‘warping consistency’ of feature maps. More specifically, SCI measures the feature similarity between the warped target frame and the reference frame without any ground truth. This approach allows the model to self-assess the quality of flow estimates in an iteration, and then also allows the model to self-correct the errors in flow estimates over iterations.

Figure~\ref{fig:sci} illustrates the concept of SCI, beginning with the input of an image pair and culminating in the output of the SCI map, which represents the ‘self-assessed quality’ of flow estimates.

Next, in the second half of this section, we elaborate on how this SCI map is derived and applied.

Given input images $I_1$ and $I_2$, we first encode these images into feature maps $F_{1,0}$ and $F_{2,0}$. We then adopt an iterative process to estimate the dense flow field $f_{1,2,i}$ in an iteration i for the dense pixelwise displacements between $F_{1,i}$ and $F_{2,i}$.
\begin{equation} \label{eq:warping}
F_{2,i}' = W(F_{2,i}, f_{1,2,i})
\end{equation}
where, W() is the standard warping operation that takes a dense input feature map $F_{2,i}$ along with the estimated dense optical flow field $f_{1,2,i}$ to produce a dense output feature map $F_{2,i}'$ by reverting the pixel-wise displacements of $F_{2,i}$ according to flow filed $f_{1,2,i}$ for each desired output coordinate point $p_{x,y}$ of $F_{1,i}$ and by interpolating closest neighboring points for the queried coordinates in the source feature map $F_{2,i}$. Taking the pointwise differences between the original $F_1$ and the warped $F_2 '$, we then apply the sum of squared differences to a Gaussian kernel function with suitable normalization as follows.
\begin{equation} \label{eq:gaussian}
G_{sci}(F_{1}, F_{2}')|_{(x,y)} = e^{\frac{-1}{2\sqrt{d}} \cdot \sum_{c}^{}(F_{1,(x,y)}-F'_{2,(x,y)})^2}
\end{equation}
where, C stands for the set of elements in the channel dimension over the corresponding coordinates ($x,y$) of $F_1$ $F_2 '$. The Gaussian kernel function comes with the following property for its value range. 
\begin{equation} \label{eq:gaussian_range}
0 \le G_{sci}(F_1, F_{2}')|_{(x,y)} \le  1,
\end{equation}
where, the maximum holds for $||F_{1,(x,y)}-F_{2,(x,y)}'||_{2}^2 =0$ and minimum holds for $||F_{1,(x,y)}-F_{2,(x,y)}'||_{2}^2 = +\infty$. For conciseness, we refer to this derived dense map $G_{sci}(F_1, F_{2}')$ as the \textit{SCI map}. We then concatenate the SCI map with the estimated dense flow map along the channel dimension and feed them as the input to ConvGRU module for iterative refinement to derive the flow adjustment on top of the estimated flow.



\subsection{Regression Focal Loss} 
\vspace{-3pt}
In this subsection, we introduce Regression Focal Loss (RFL). Given the observation that the difficulty in predicting the pixel-wise flow can differ from one pixel to another depending on the contents at and around the pixels, we aim at helping the network focus its learning on regions that needs more improvement. Comparing with the focal loss \cite{lin2017focal} used in segmentation for handling class imbalance, our proposed RFL is intended for dense regression instead and may be considered as for "difficulty imbalance". To this end, we first derive a confidence map to facilitate the pixel-wise weighting. We adopt the confidence map in LiteFlowNetv3 \cite{hui2020liteflownet3} as follows. 
\begin{equation} \label{eq:conf}
M_\text{conf}(x) = e^{-||f_\text{gt}(x) - f_\text{pred}(x)||^2}
\end{equation}

Having the confidence map ready, we apply the map to $l_\text{i}$ and replace Eq.~\ref{eq:optical_each_l1} with the following. 
\begin{equation} \label{eq:rfl_l1}
l_\text{i} = ||(1 + \alpha \cdot (1-M)^{\beta}) \cdot (f_\text{gt} - f_\text{i})||_1
\end{equation}
where $\alpha$ and $\beta$ are hyper parameters. The intuition of Eq. \ref{eq:rfl_l1} is that we apply higher weighting to regions of low confidence and standard weighting to high confidence regions.

This RFL-based confidence weighting is derived by the final iteration of prediction and applied $l_\text{i}$ of all iterations.
We find this to be more effective as confidence derived in earlier iterations tends to be noisier, as we shall discuss more in our ablation study. Figure \ref{fig:rfl} compares between the baseline and the RFL approaches.

\begin{table*}[t]
\begin{center}
\caption{Ablation study for Self-Cleaning Iteration (SCI) and Regression Focal Loss (RFL). Following same protocol as described in RAFT \cite{teed2020raft}, we train all model variants on top of two baseline architectures by two combinations of datasets specified in the table and evaluate them on Sintel (S) and KITTI (K) training datasets. 
}
\vspace{-8pt}
\label{tab:csirfl_ablation}
\adjustbox{max width=0.8\textwidth}
{
\begin{tabular}{|l|c||l|c|c||c|c||c|c|}
\hline
\multirow{2}{*}{} & Training & \multirow{2}{*}{Architecture} & \multirow{2}{*}{SCI} & \multirow{2}{*}{RFL} &  \multicolumn{2}{|c|}{Sintel}  & \multicolumn{2}{|c|}{KITTI 15} \\
\cline{6-9}
& Datasets & & & & clean (epe) & final (epe) & Fl-epe & Fl-all\\ 
\hline
\hline
\multirow{7}{*}{\rotatebox{90}{\textbf{Cross-Domain}}} & \multirow{7}{*}{C+T} &\multirow{4}{*}{RAFT-small \cite{teed2020raft}}  & & & 2.21 (-0.0\%) & 3.35 (-0.0\%) & 7.51 (-0.0\%) & 26.90 (-0.0\%) \\
&&&  & $\checkmark$ & 2.29 \ (+3.6\%) & 3.52 \ (+5.1\%) & 7.44 (-0.9\%) & 24.88 (-7.5\%) \\
&&& $\checkmark$ & & 2.17  (-1.8\%) & \textbf{3.33  (-0.6\%)} & 7.58  (-0.9\%) & 25.45  (-5.4\%)\\
&&& $\checkmark$ & $\checkmark$ & \textbf{2.11  (-4.5\%)} & 3.34  (-0.3\%) & \textbf{7.22  (-3.9\%)} & \textbf{24.62  (-9.5\%)}\\
\cline{3-9}
&&\multirow{3}{*}{MobileFlow\textsuperscript{\ref{mobileflow}}}  & & & 1.79 (-0.0\%) & 3.47 (-0.0\%) & 8.33 (-0.0\%) & 22.06 (-0.0\%) \\
&&& $\checkmark$ & & \textbf{1.65  (-7.8\%)} & \textbf{3.30 (-4.9\%)} & 7.22 (-13.3\%) & \textbf{20.73  (-6.0\%)}\\
&&& $\checkmark$ & $\checkmark$ & 1.68 (-6.1\%) & 3.34  (-3.7\%) & \textbf{7.21  (-13.4\%)} & 20.75 (-5.9\%) \\
\hline
\hline
\multirow{6}{*}{\rotatebox{90}{\textbf{In-Domain}}} & \multirow{6}{*}{C+T + S/K} &\multirow{3}{*}{RAFT-small \cite{teed2020raft}}  & & & \textbf{1.42 (-0.0\%)} & 2.09 (-0.0\%) & 1.21 (-0.0\%) & 4.68 (-0.0\%) \\
&&& $\checkmark$ & & 1.46  \ (+2.8\%) & 2.06 (-1.4\%) & \textbf{1.20 (-0.8\%)} & 4.74 (-1.3\%) \\
&&& $\checkmark$ & $\checkmark$ & 1.46 \ (+2.8\%) & \textbf{2.04 (-2.4\%)} & \textbf{1.20 (-0.8\%)} & \textbf{4.50 (-3.8\%)} \\
\cline{3-9}
&&\multirow{3}{*}{MobileFlow\textsuperscript{\ref{mobileflow}}}  & & & 1.09 (-0.0\%) & 1.76 (-0.0\%) & 0.96 (-0.0\%) & 3.14 (-0.0\%) \\
&&& $\checkmark$ & & 1.09 (-0.0\%) & 1.74 (-1.1\%) & 0.94 (-2.1\%) & 2.97 (-5.4\%) \\
&&& $\checkmark$ & $\checkmark$ & \textbf{1.03 (-5.5\%)} & \textbf{1.65 (-6.3\%)} & \textbf{0.92 (-4.2\%)} & \textbf{2.81 (-10.5\%)} \\
\hline
\end{tabular}
}
\vspace{-5mm}
\vspace{2pt}
\end{center}
\end{table*}

\begin{table*}[t]
\begin{center}
\caption{Ablation study of Regression Focal Loss (Eq.~\ref{eq:rfl_l1}). We train RAFT-small+SCI models on FlyingChairs (C) and FlyingThings (T) and evaluate on Sintel (S) and KITTI (K) training datasets. 
}
\vspace{-8pt}
\label{tab:constant1_ablation}
\adjustbox{max width=0.7\textwidth}
{
\begin{tabular}{|l||c|c||c|c|}
\hline
\multirow{2}{*}{$l_i$ loss}  &  \multicolumn{2}{|c|}{Sintel}  & \multicolumn{2}{|c|}{KITTI 15} \\
\cline{2-5}
 & clean (epe) & final(epe)  & Fl-epe & Fl-all\\ 
\hline
\hline
$a. \thickspace\thickspace ||(f_\text{gt} - f_\text{i})||_1$  (Eq.~\ref{eq:optical_each_l1}) & 2.17 (-0.0\%) & \textbf{3.33 (-0.0\%)}  & 7.58 (-0.0\%) & 25.45 (-0.0\%) \\
$b. \thickspace\thickspace ||(\alpha \cdot (1-M)^{\beta}) \cdot (f_\text{gt} - f_\text{i})||_1$ & 2.32 (+6.9\%) & 3.56 (+6.9\%) & \textbf{7.09 (-6.5\%)} & 25.27 (-0.7\%) \\
$c. \thickspace\thickspace ||(1 + \alpha \cdot (M)^{\beta}) \cdot (f_\text{gt} - f_\text{i})||_1$  & 4.52 (+108.3\%) & 5.92 (+77.8\%) & 10.07 (+32.8\%) & 50.53 (+95.5\%) \\
$d. \thickspace\thickspace ||(1 + \alpha \cdot (1-M)^{\beta}) \cdot (f_\text{gt} - f_\text{i})||_1$ (Eq.~\ref{eq:rfl_l1}) & \textbf{2.11 (-2.8\%)} & 3.34 (+0.3\%) & 7.22 (-4.7\%) & \textbf{24.62 (-3.3\%)}\\
\hline
\end{tabular}
}
\vspace{-5mm}
\end{center}
\end{table*}

\subsection{SciFlow: The Combination of SCI and RFL}
\vspace{-3pt}

Having individual definitions for SCI and RFL, we further discuss their relationship and our final proposal for combining them. Despite that SCI in Eq.~\ref{eq:gaussian} and RFL in Eq.~\ref{eq:conf} share similar arithmetic structures, their sources of the feature maps for the contrastive measures are quite different. During training, while the RFL relies on the ground truth in back propagation to focus on regions of larger residual regression errors, the SCI relies completely on the input images in the forward pass to derive the SCI map. During inference, the model continues its active computation for the SCI map to self-assess the flow estimates and to self-clean the flow ambiguities. SCI and RFL seem to be synergistic in learning to handle feature ambiguities, while they also complement each other in how their contrastive measures are used. In Section \ref{sec:experiments}, we discuss more on empirical results for the combination of SCI and RFL.


%% file: sec/4_experiments.tex

\section{Experiments}
\label{sec:experiments}

\subsection{Experimental Setup}
\vspace{-3pt}
\textbf{Datasets:}
We follow commonly adopted training and evaluation protocols in the literature~\cite{teed2020raft, jiang2021learning, zhang2021separable, huang2022flowformer}. We train our model on FlyingChairs (C)~\cite{dosovitskiy2015flownet} and FlyingThings3D (T)~\cite{mayer2016large} and evaluate on training dataset of Sintel (S)~\cite{butler2012naturalistic} and KITTI (K)~\cite{geiger2013vision, menze2015object, Menze2015ISA}. Using C+T pre-trained model, we finetune Sintel and KITTI datasets and evaluate on Sintel and KITTI datasets. 


\textbf{Network Architectures and Training:}
\label{subsec:nwkarch}
We use two lightweight models with different architectures, RAFT-small~\cite{teed2020raft} and MobileFlow\footnote{\label{mobileflow}MobileFlow is our created lightweight baseline architecture. Please see section \ref{subsec:nwkarch} "Network Architectures and Training" for more details.}, as our baselines in the experiments. In particular, MobileFlow is our model creation for a lightweight baseline architecture. In order to build a feasible architecture that fits within the limited memory and compute capacity of a smartphone, we utilize memory-efficient cost volume techniques from~\cite{xu2021high, zhang2021separable, Jiang2021LearningOF}. We also adopt a MobileNetV2~\cite{sandler2018mobilenetv2} based backbone for feature extraction and a ConvGRU module for iterative refinement that is similar to~\cite{teed2020raft}. 
For fair comparisons, we train both RAFT-small and MobileFlow baselines along with all their variants for SCI and RFL on top of the baselines using same train framework\footnote{RAFT: \url{https://github.com/princeton-vl/RAFT}} and dataset protocol as described in RAFT \cite{teed2020raft} to report our experiment results. We follow the training parameters all the same as for RAFT, including number of iterations and the learning rate. For additional parameters of regression focal loss, we set both $\alpha$ and $\beta$ in Eq.~\ref{eq:rfl_l1} to $1$.

\textbf{Evaluation Metrics:} We evaluate our models by the End-Point Error (EPE) metric, which is the Euclidean distance between the predicted flow and the ground truth flow. We also use F1-all as defined for the KITTI dataset \cite{Menze2015ISA}. In both cases of error metrics, the lower is the better. 


\vspace{4pt}
\subsection{Experimental Results}
\vspace{-3pt}

\subsubsection{Cross-Domain Evaluation}
\vspace{-2pt}
The top half of Table~\ref{tab:main_table} shows our cross-domain evaluation results, for which the models are trained on FlyingChairs and FlyingThings, and then are evaluated on Sintel and KITTI training datasets, respectively. Our solution, MobileFlow+SciFlow (namely, with both SCI and RFL), achieves significantly higher accuracy not only over the baseline MobileFlow but also over other compared lightweight optical flow methods.


\begin{figure*}[h]
\begin{center}$
\centering
\begin{tabular}{c c c c }
\hspace{-0.3cm} \rotatebox{90}{\textbf{\ \ \quad Image 1}} & \hspace{-0.4cm} \includegraphics[width=5.6cm]{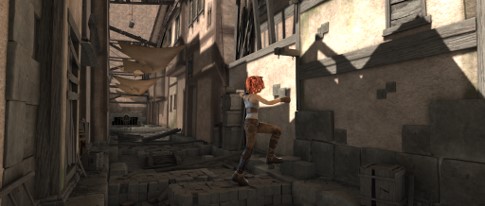} 
& \hspace{-0.4cm} \includegraphics[width=5.6cm]{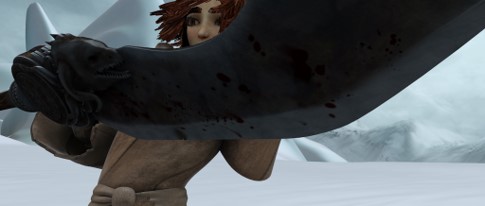} 
& \hspace{-0.4cm} \includegraphics[width=5.6cm]{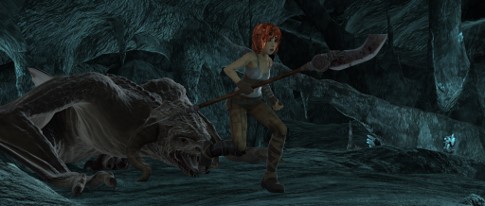} \\
 \hspace{-0.3cm} \rotatebox{90}{\textbf{\ \ \quad Image 2}} & \hspace{-0.4cm} \includegraphics[width=5.6cm]{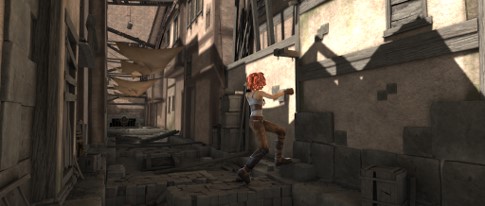} 
& \hspace{-0.4cm} \includegraphics[width=5.6cm]{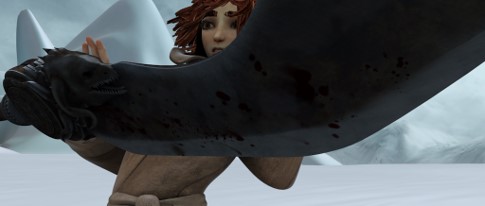} 
& \hspace{-0.4cm} \includegraphics[width=5.6cm]{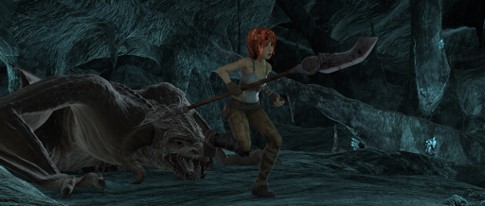} \\
\hspace{-0.3cm} \rotatebox{90}{\textbf{\ \ \small{Ground Truth}}} & \hspace{-0.4cm} \includegraphics[width=5.6cm]{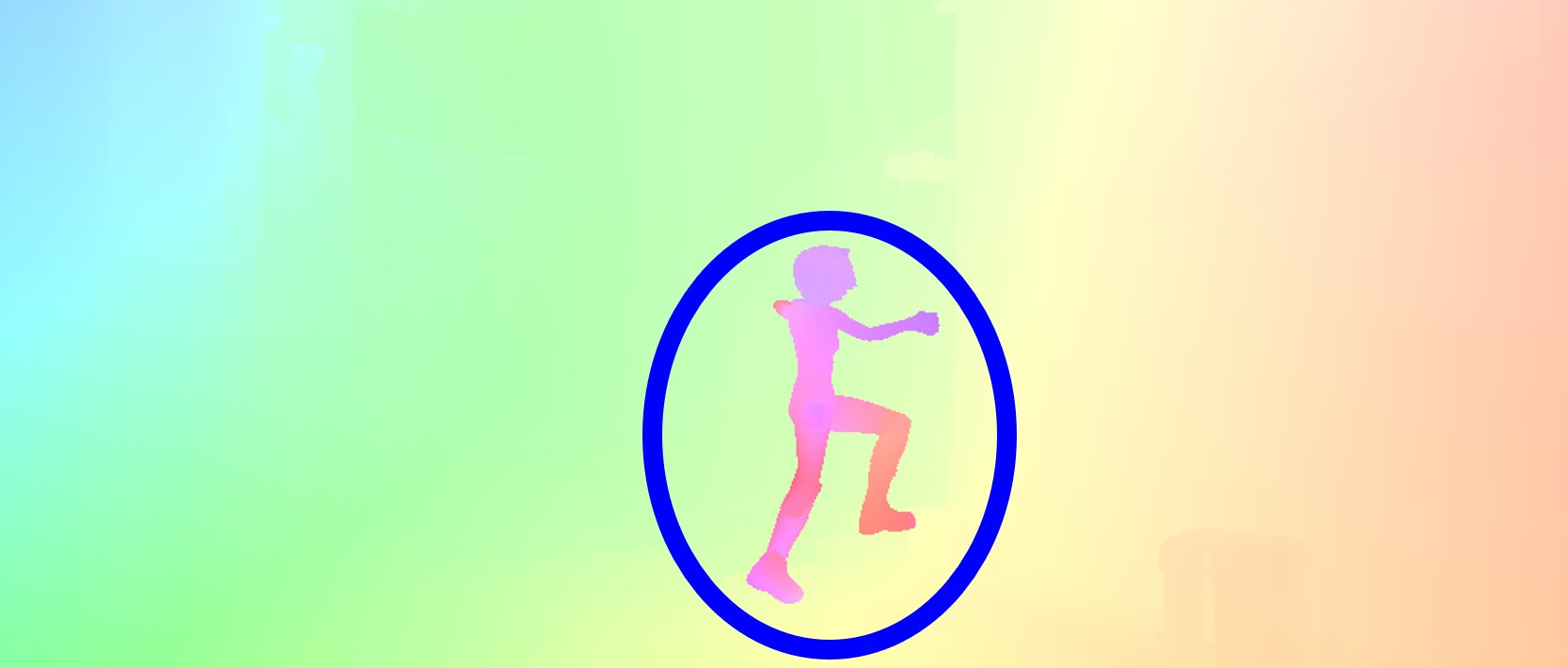} 
& \hspace{-0.4cm} \includegraphics[width=5.6cm]{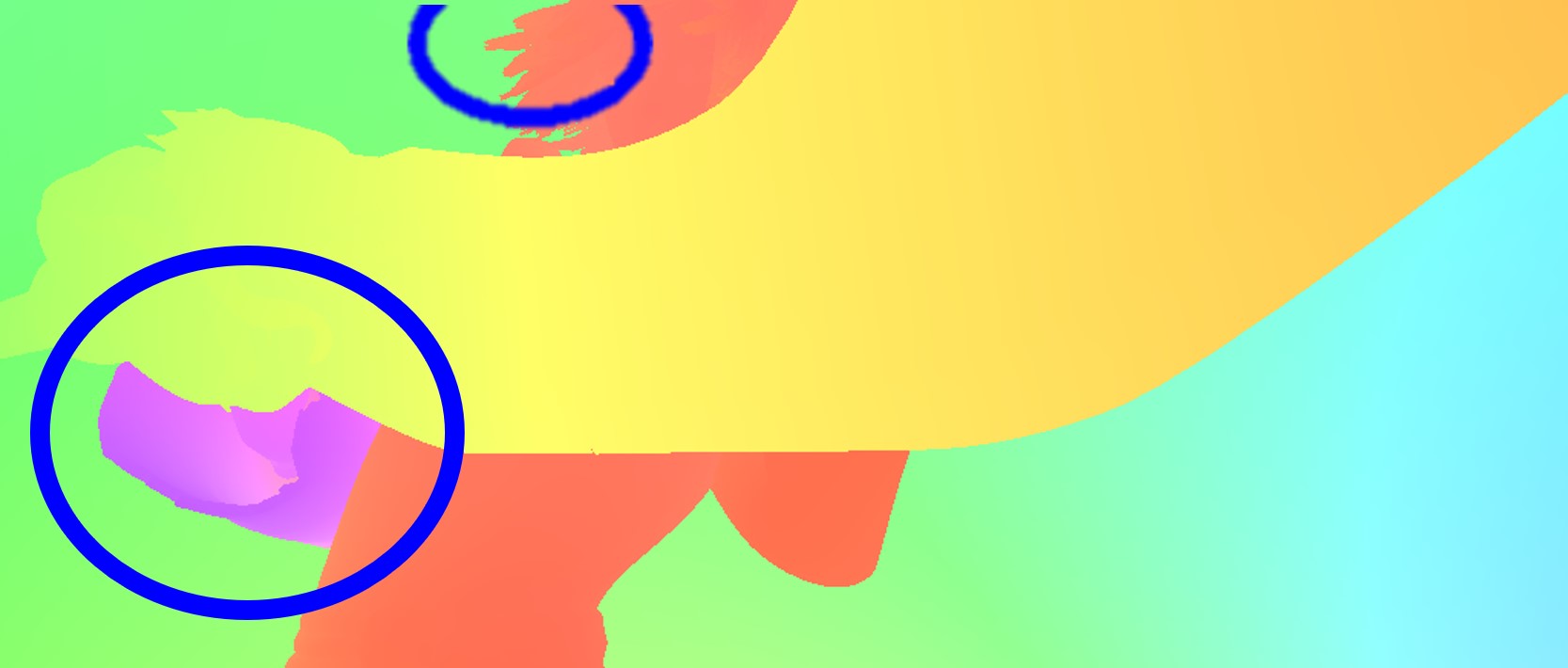} 
& \hspace{-0.4cm} \includegraphics[width=5.6cm]{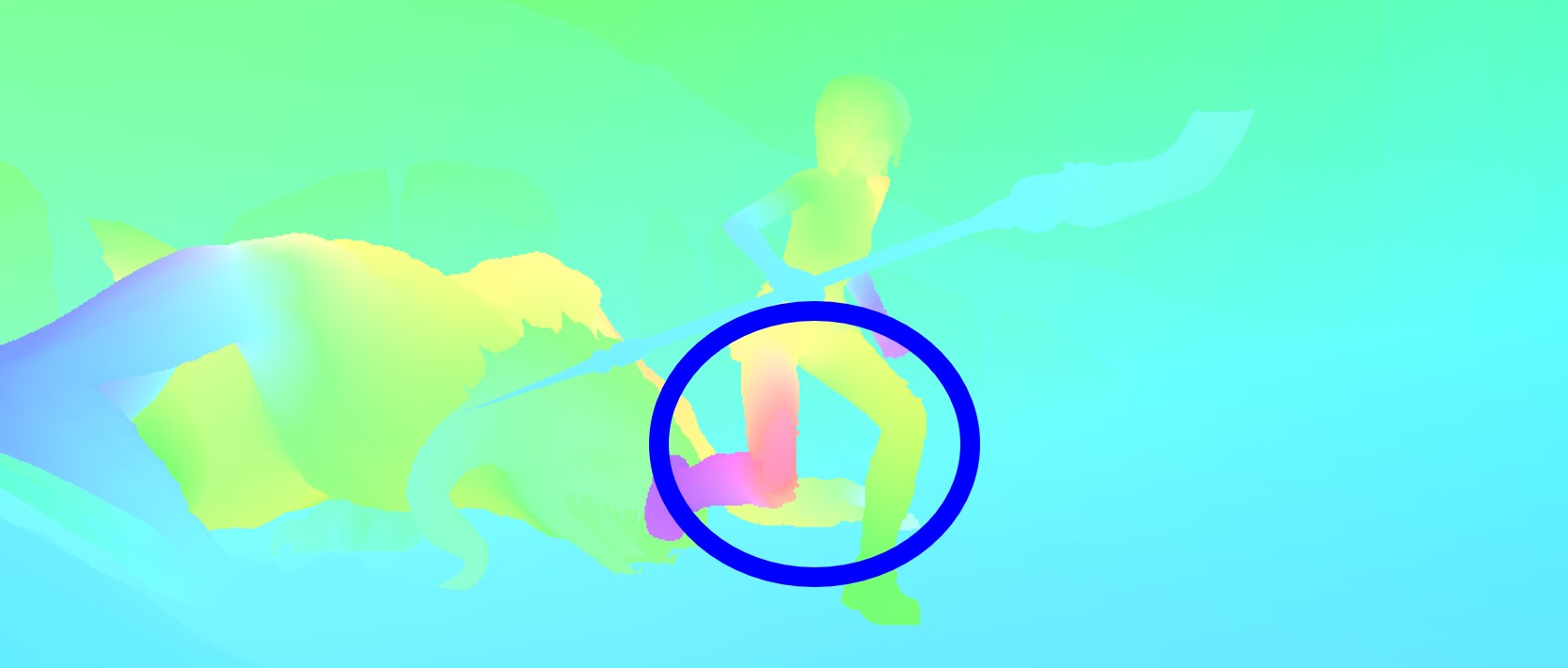} \\
\hspace{-0.3cm} \rotatebox{90}{\textbf{\ \quad Baseline}} & \hspace{-0.4cm}  \includegraphics[width=5.6cm]{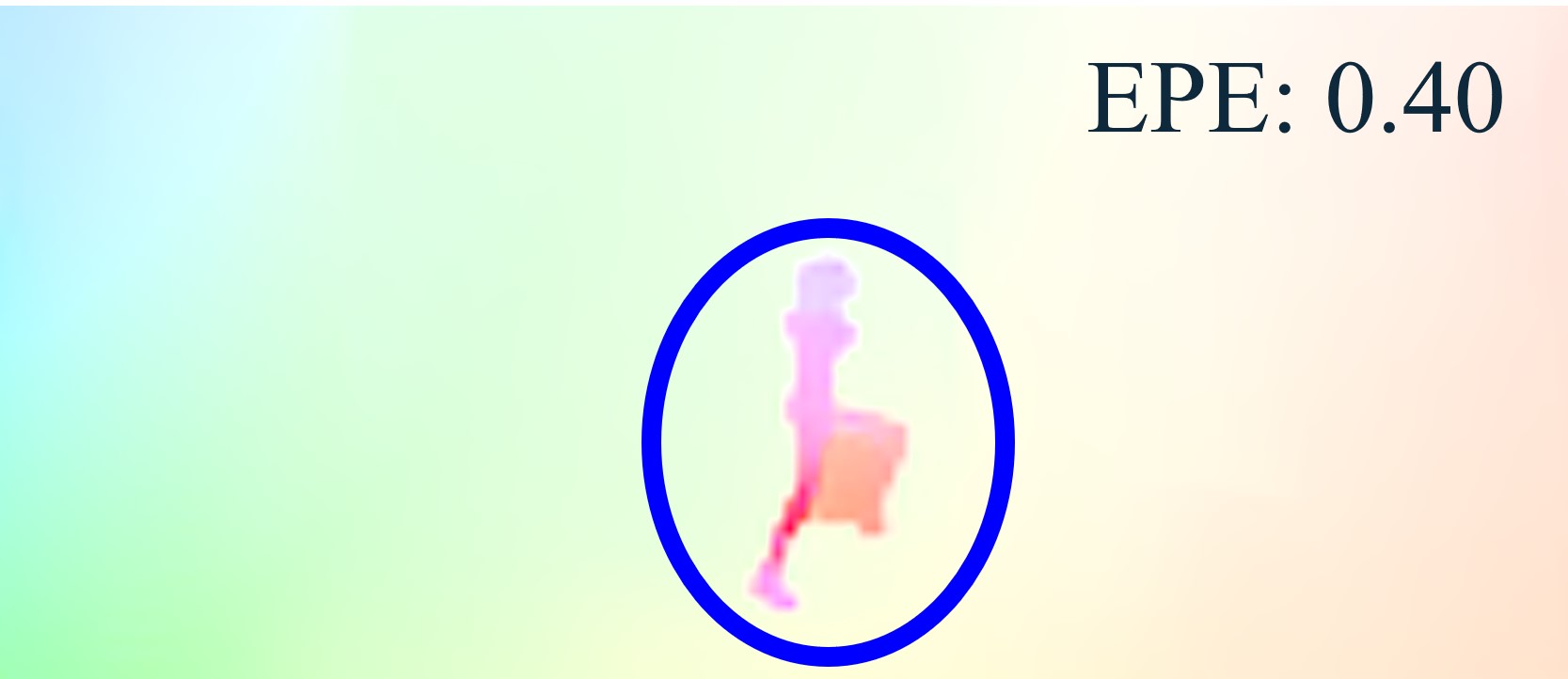} 
& \hspace{-0.4cm} \includegraphics[width=5.6cm]{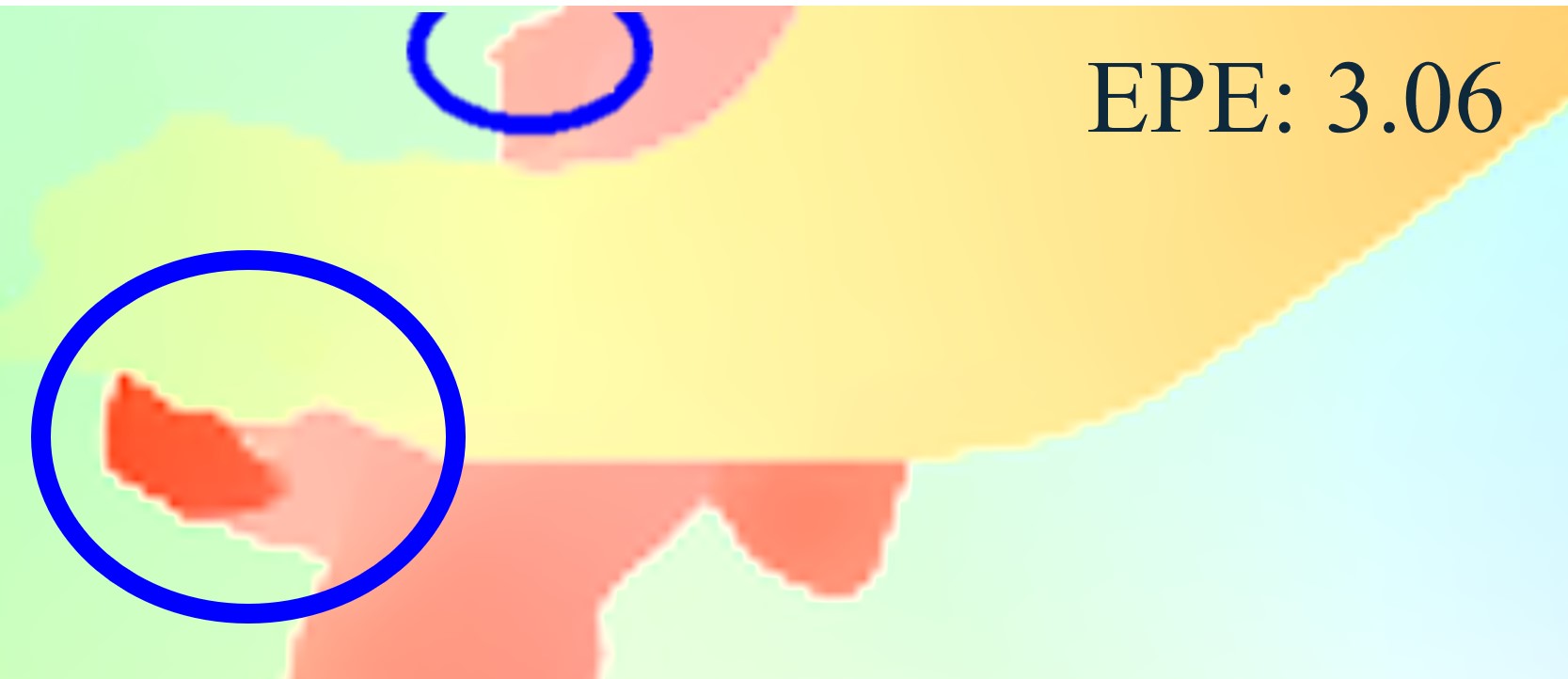} 
& \hspace{-0.4cm} \includegraphics[width=5.6cm]{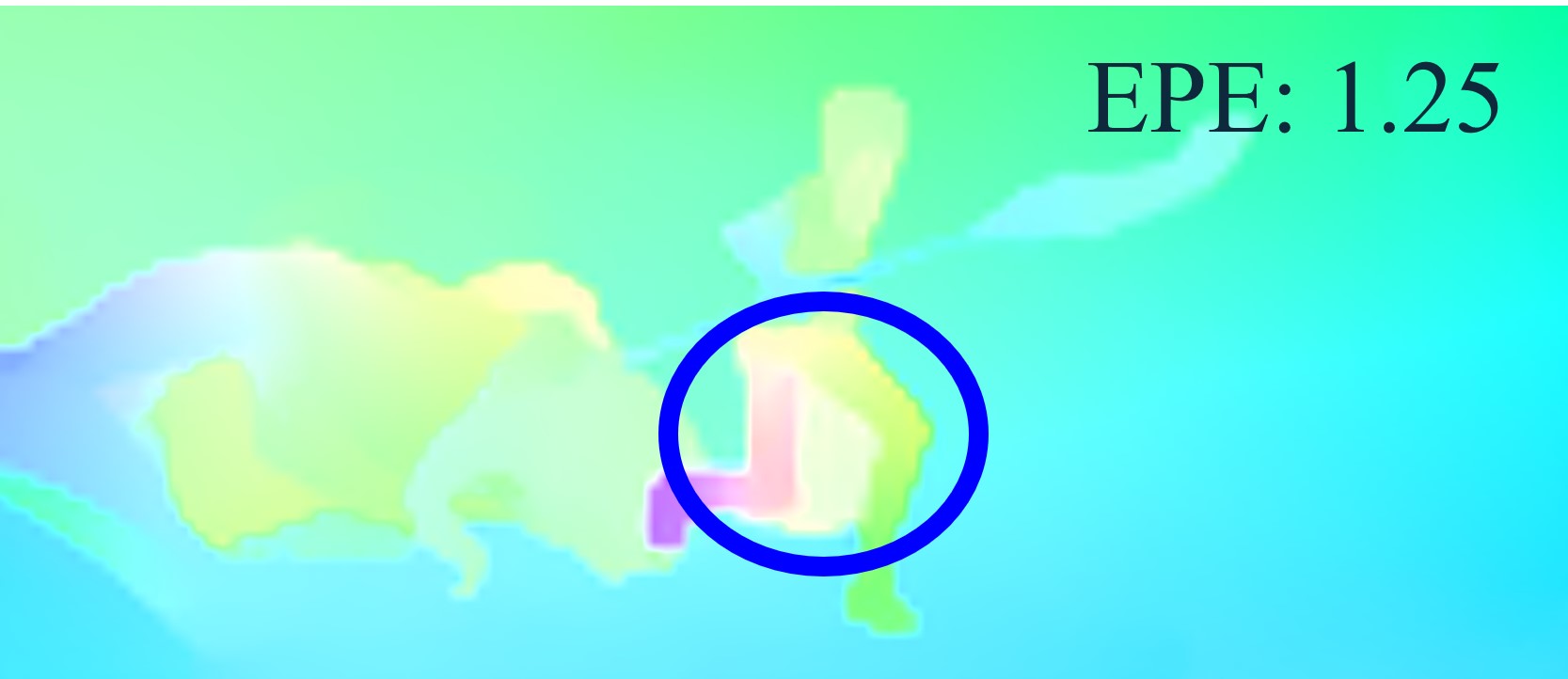} \\
 \hspace{-0.3cm} \rotatebox{90}{\textbf{ \quad Base + SCI}} & \hspace{-0.4cm}  \includegraphics[width=5.6cm]{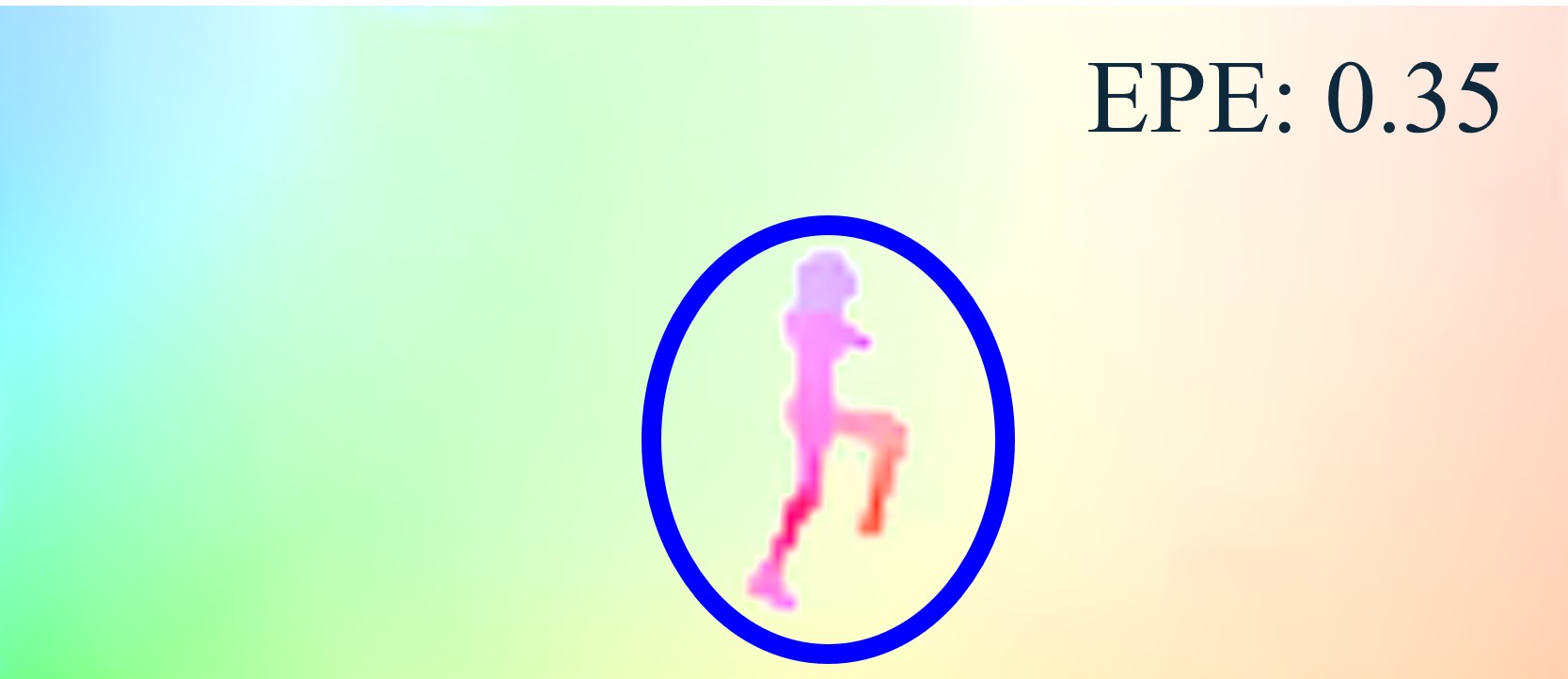} 
& \hspace{-0.4cm} \includegraphics[width=5.6cm]{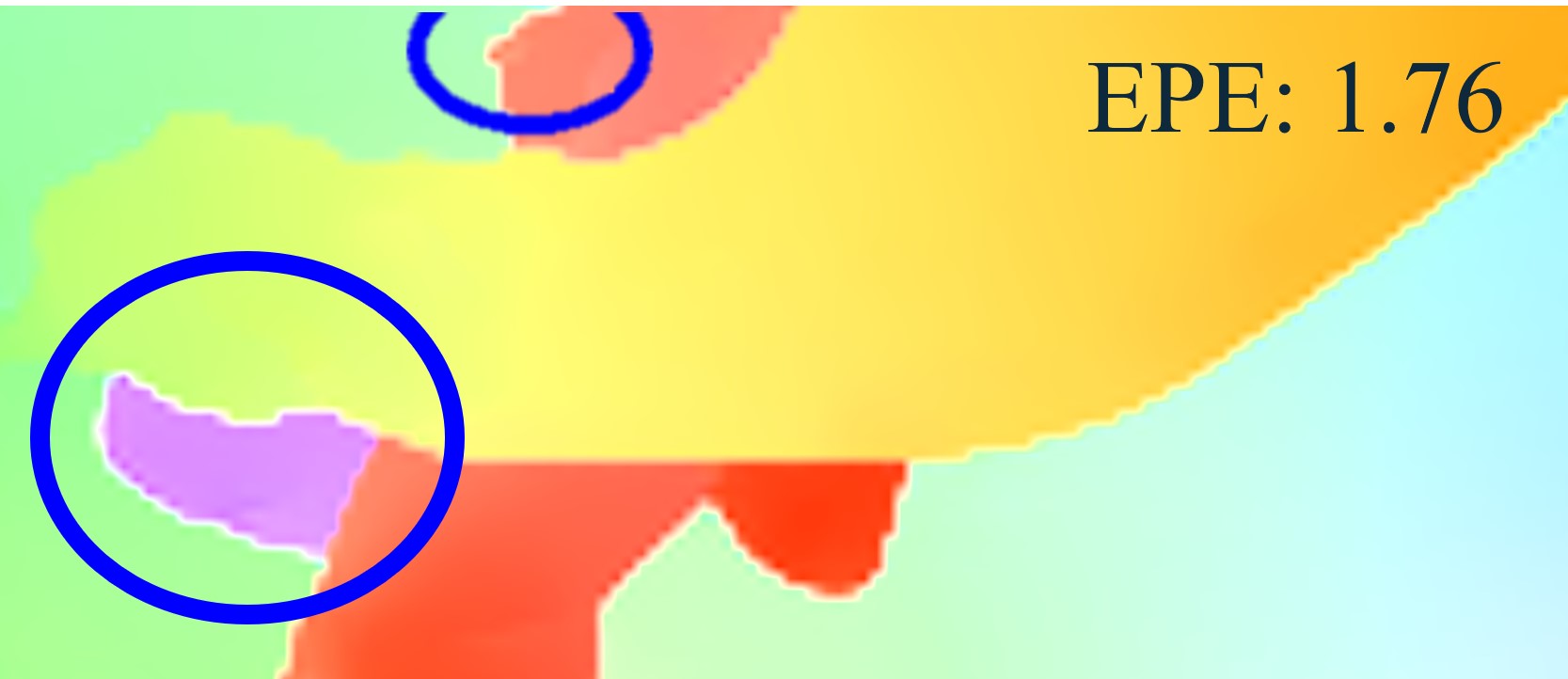} 
& \hspace{-0.4cm} \includegraphics[width=5.6cm]{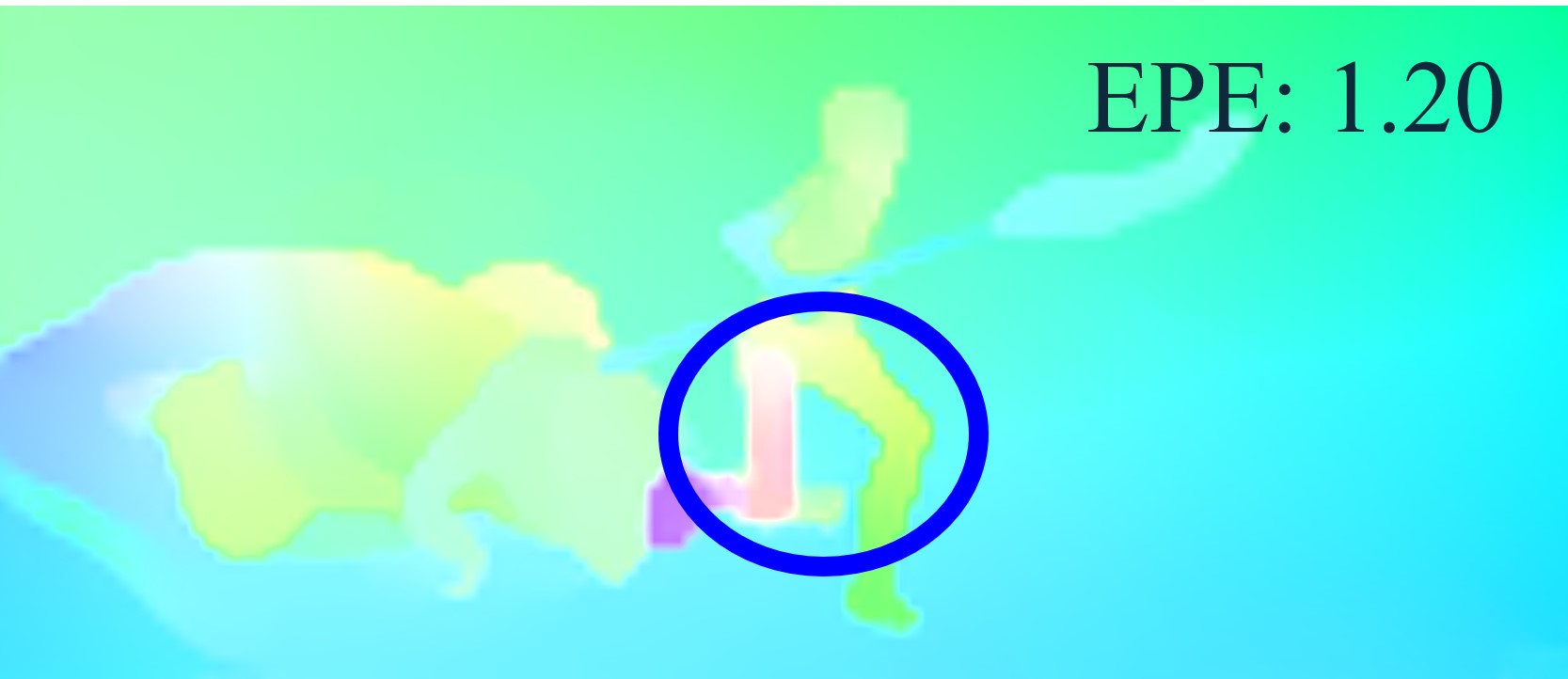} \\
 \hspace{-0.3cm} \rotatebox{90}{\textbf{\small{Base + SCI+ RFL}}} & \hspace{-0.4cm} \includegraphics[width=5.6cm]{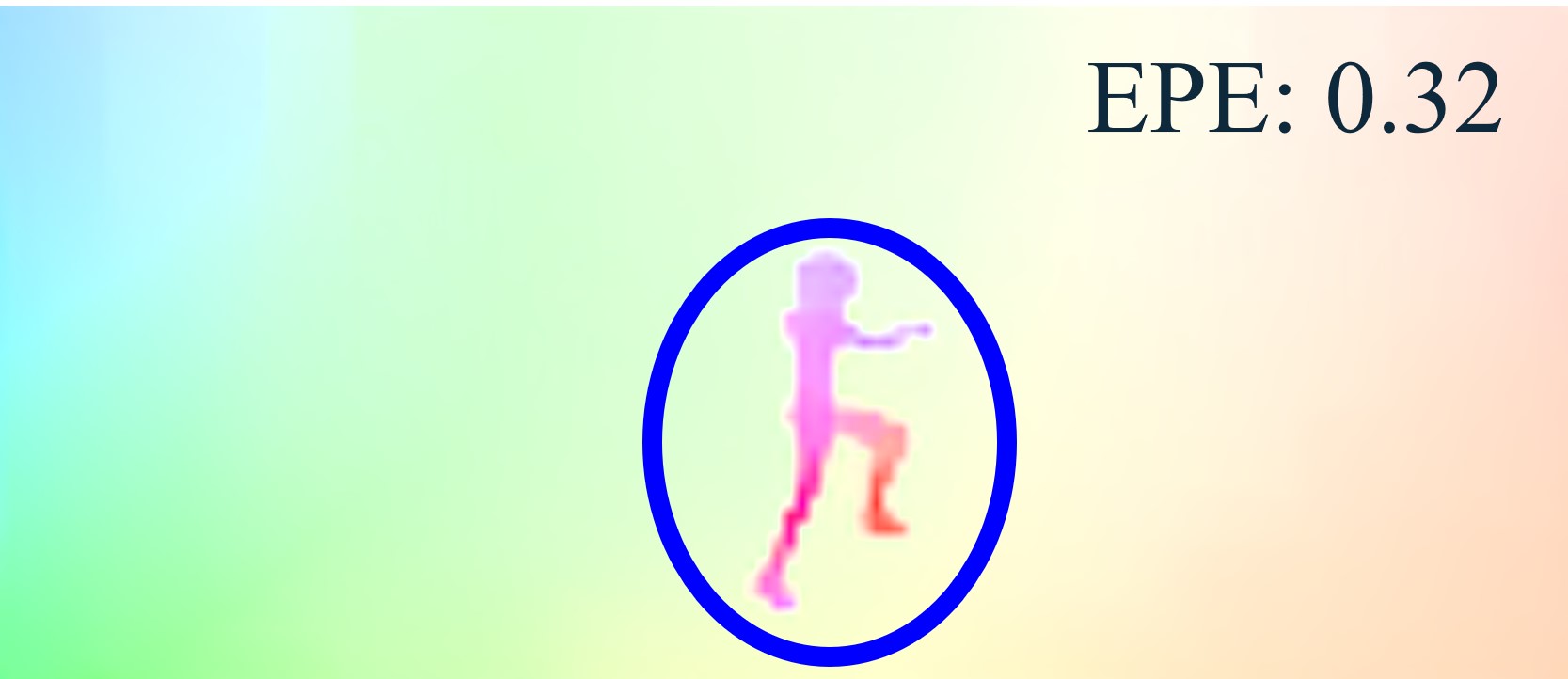} 
& \hspace{-0.4cm} \includegraphics[width=5.6cm]{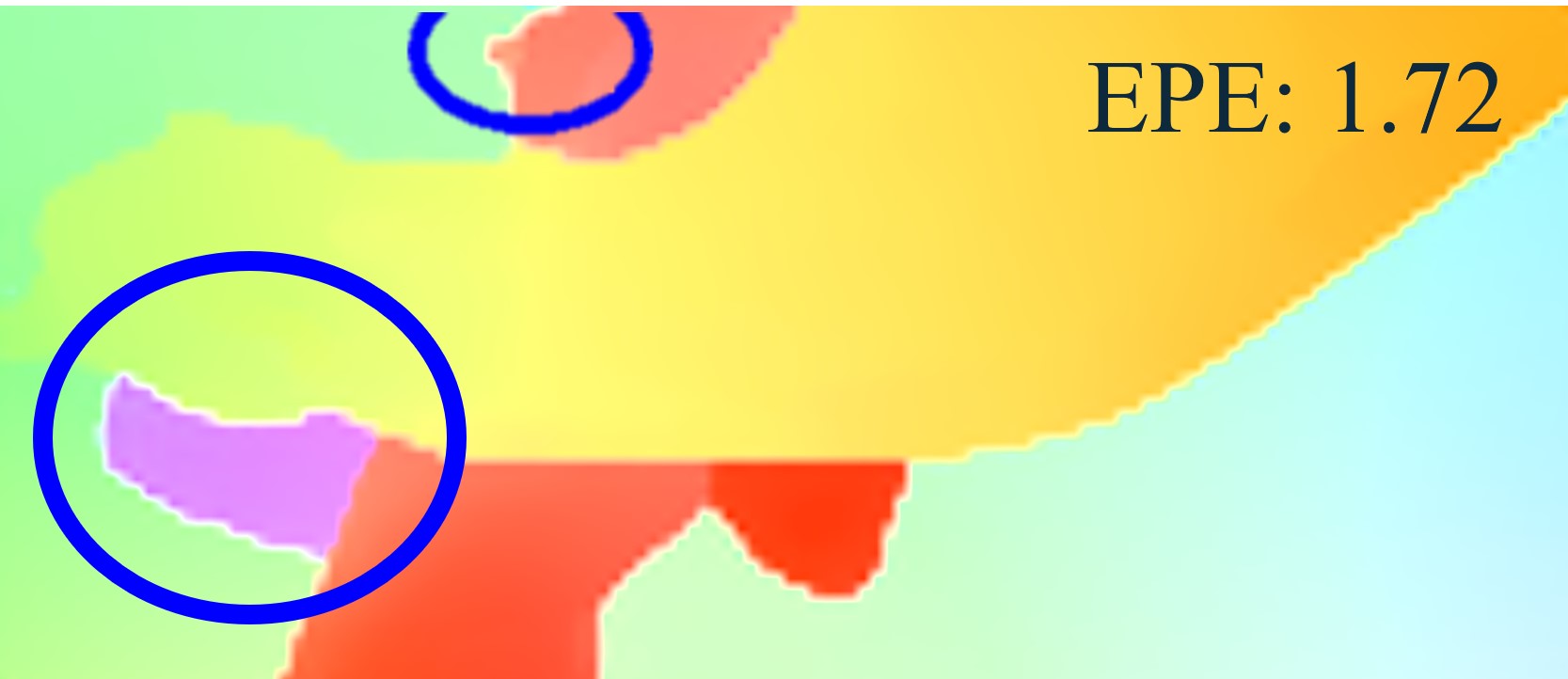} 
& \hspace{-0.4cm} \includegraphics[width=5.6cm]{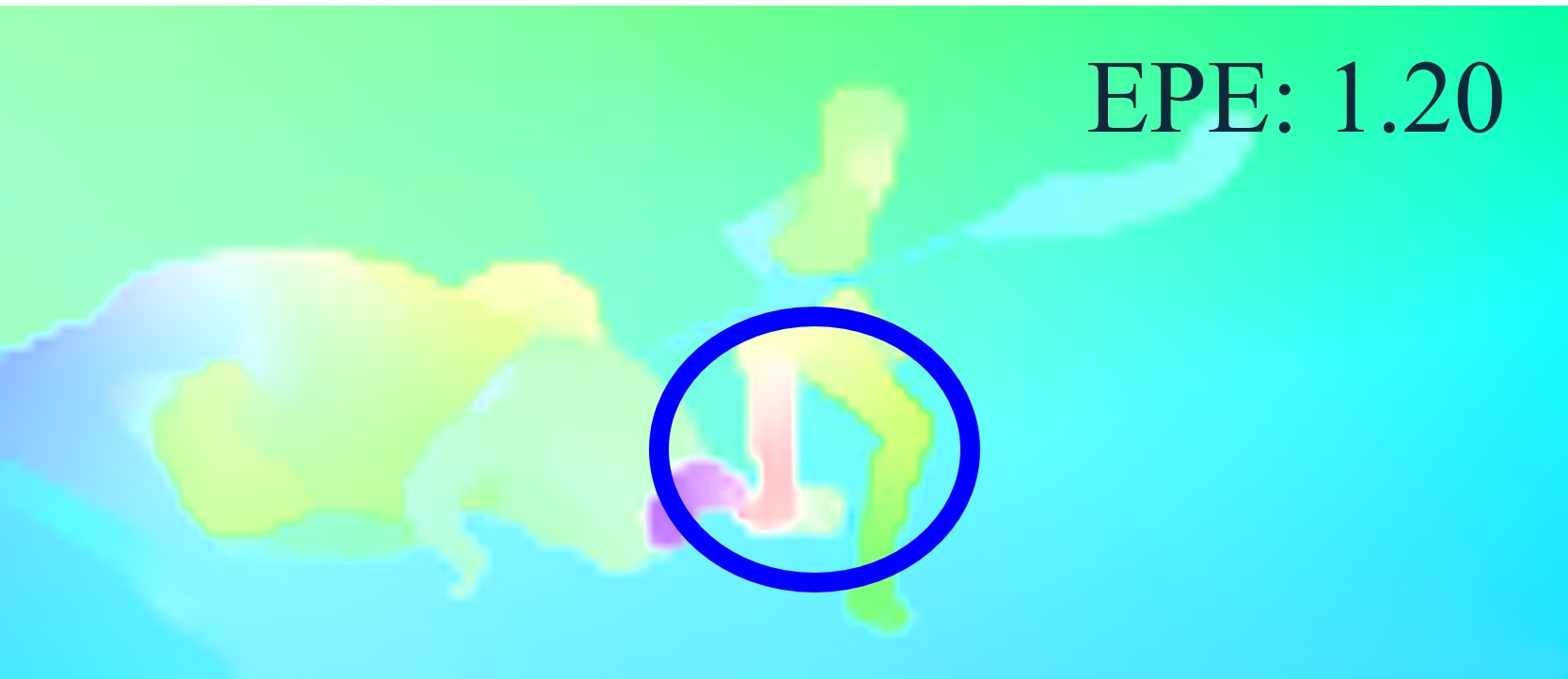} 
\end{tabular}$
\end{center}
\vspace{-18pt}
\caption{
\textbf{Qualitative results on Sintel (train) dataset using RAFT-small architecture (trained with C+T).} First and second rows are input images. Third row is the ground truth. Fourth row is the output of RAFT-small. Fifth and sixth rows are the output of RAFT-small + SCI and RAFT-small + SCI + RFL, respectively. 
}
\label{exp:qualitative_sintel}
\vspace{-4mm}
\vspace{3pt}
\end{figure*}

\vspace{-2pt}
\subsubsection{In-Domain Evaluation}
\vspace{-2pt}
The bottom half of Table~\ref{tab:main_table} shows our in-domain evaluation results, where models are trained on FlyingChairs, FlyingThings3D, and Sintel (or KITTI) and are evaluated on Sintel (or KITTI) following the protocol as in RAFT \cite{teed2020raft}. Our proposed solution demonstrates significantly improved accuracy over the baseline and even over other state-of-the-art lightweight optical flow models. Please note that, unlike LiteFlowNet model series, in our experiments MobileFlow and its variant are trained only on KITTI 2015 dataset but not also on KITTI 2012 dataset.

\begin{figure*}[h]
\begin{center}$
\centering
\begin{tabular}{c c c c}
 & Baseline & Base+SCI & Base+SCI+RFL \\
\vspace{-0.1cm} 
\hspace{-0.4cm} \rotatebox{90}{\textbf{\ \footnotesize{Confidence Maps}}} & \hspace{-0.4cm} \includegraphics[width=5.7cm]{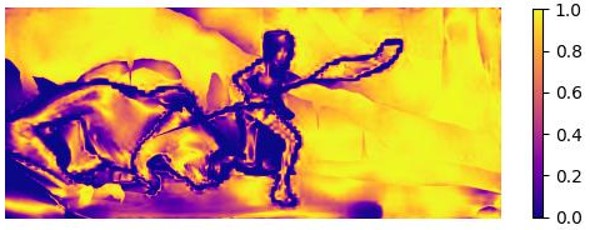} 
& \hspace{-0.5cm} \includegraphics[width=5.7cm]{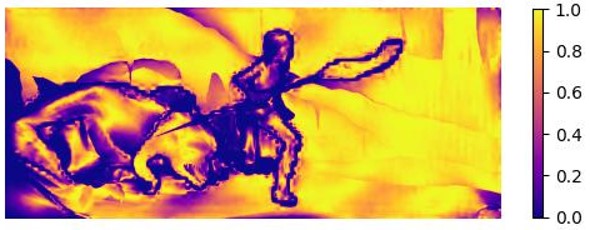} 
& \hspace{-0.5cm} \includegraphics[width=5.7cm]{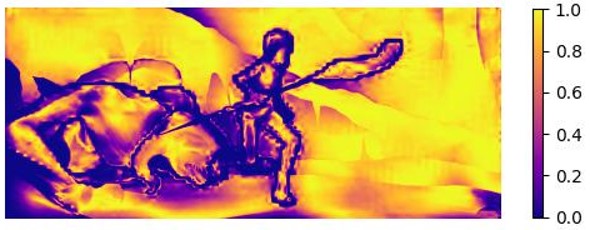} \\
\vspace{-0.3cm} 
\hspace{-0.4cm} 
\rotatebox{90}{\textbf{\quad \small{Error Maps}}} & \hspace{-0.4cm} \includegraphics[width=5.7cm]{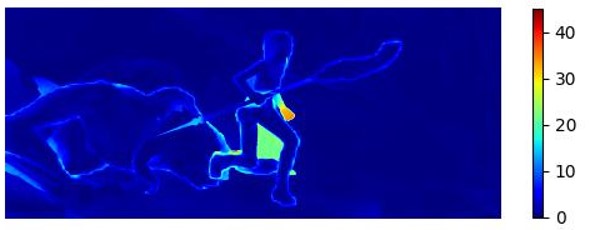} 
& \hspace{-0.5cm} \includegraphics[width=5.7cm]{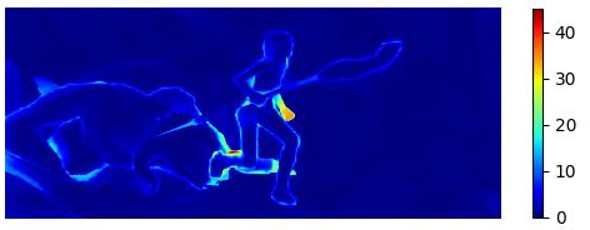} 
& \hspace{-0.5cm} \includegraphics[width=5.7cm]{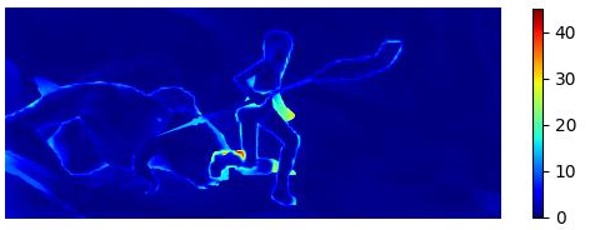} \\
\end{tabular}$
\end{center}
\vspace{-10pt}
\caption{\textbf{Subtleties in RFL-based confidence maps and error maps.} \textbf{Top Row:} We report confidence maps derived with our RFL technique for the baseline and its variants. Moreover, SCI and RFL demonstrate their abilities to help resolve ambiguities in certain regions, as indicated by their higher confidence measures in these maps. \textbf{Bottom Row:} We report error maps for the predicted flow estimates against the ground truth for the baseline and its variants. Moreover, SCI and RFL demonstrate their ability to help resolve ambiguities in certain regions, as evidenced by their lower errors in these maps.
In these examples, we use samples from the Sintel (train) dataset with RAFT-small architecture. The original input images can be found in the right column of Fig~\ref{exp:qualitative_sintel}.
}
\label{exp:confidence_qualitative_sintel}
\vspace{-3mm}
\end{figure*}


\subsubsection{Ablation Study}
\textbf{SCI vs. RFL:} Table~\ref{tab:csirfl_ablation} summarizes our ablation study over choices and/or combinations of SCI and RFL. We applied SCI and RFL into three variants based on either the architecture of RAFT-small or the MobileFlow. Despite the fact that not all variants demonstrate improved accuracy in these experiments, the particular variants base+SCI+RFL in general demonstrate competitive accuracy over their respective baselines among minor run-to-run variations.

\textbf{Regression Focal Loss:} 
Table~\ref{tab:constant1_ablation} summarizes our ablation study on RFL (Eq~\ref{eq:rfl_l1}). Option "a" is the standard L1 loss without applying RFL. Option "b" produces inconsistent results over Sintel and KITTI, suggesting the impact of removing the portion of the standard L1 loss. Option "c" uses the opposite focus on the regions of high confidence, which interestingly produces drastic degradation in accuracy, suggesting the wrong focus for the learning. Our proposed form in option "d" produces competitive results by combing both the L1 loss and the confidence-weighted focus on regions of higher residual errors.

\begin{table}[t]
\begin{center}
\caption{Ablation study to compare between final-iteration confidence map and per-iteration confidence map. Here we train RAFT-large models on FlyingChairs (C) and FlyingThings (T) and evaluate on Sintel (S) and KITTI (K) training datasets. 
}
\vspace{-8pt}
\label{tab:confidence_ablation}
\adjustbox{max width=0.48\textwidth}
{
\begin{tabular}{|l||c|c||c|c|}
\hline
\multirow{2}{*}{Source of $M_{conf}$ (Eq.~\ref{eq:conf})}  &  \multicolumn{2}{|c|}{Sintel}  & \multicolumn{2}{|c|}{KITTI 15} \\
\cline{2-5}
 & clean (epe) & final (epe)  & Fl-epe & Fl-all\\ 
\hline
\hline
No confidence map & 1.43 (-0.0\%) & \textbf{2.71 (-0.0\%)}  & 5.04 (-0.0\%) & 17.4 (-0.0\%) \\
Per-iter confidence map & 1.41 (-1.4\%) & 2.75 \ (+1.5\%)  & 4.63 (-8.1\%) & 16.6 (-4.6\%) \\
Final-iter confidence map & \textbf{1.38 (-3.5\%)} & 2.77 \ (+2.2\%) & \textbf{4.58 (-9.1\%)} & \textbf{16.2 (-6.9\%)} \\
\hline
\end{tabular}
}
\vspace{-9mm}
\end{center}
\end{table}

\textbf{Final-Iteration Confidence Map vs. Per-Iteration Confidence Map:} 
Our proposed approach is to apply the single final confidence map to all iterations. When we apply instead per-iteration confidence map for each individual iteration, we see smaller gains than in the proposed approach. Table~\ref{tab:confidence_ablation} lists the numbers.



\subsubsection{Qualitative Results}
\vspace{-1mm}

Fig.~\ref{exp:qualitative_sintel} gives qualitative samples on Sintel dataset. The base+SCI variant demonstrates improved robustness over the baseline in occlusion areas. Base+SCI+FRL further shows slight improvements in several subtle visual details.


\vspace{-2mm}
\subsubsection{RFL as A Confidence Measure}
\vspace{-1mm}
Figure \ref{exp:confidence_qualitative_sintel} demonstrates an additional use of RFL as a confidence measure in inference.

\vspace{-2mm}

\subsubsection{On-Device Evaluation}
\label{subsect:ondevice}

\vspace{-1mm}

We report on-device evaluation of our models on Samsung S24 with a Snapdragon 8 Gen 3 processor and Qualcomm\textsuperscript{\faRegistered}
 Hexagon$^{\text{TM}}$ Tensor Processor (HTP), which is an AI accelerator specialized for neural network workloads. We adopt the INT8 (W8A8) quantization based on AIMET\footnote{AIMET is a product of Qualcomm Innovation Center, Inc.} \cite{siddegowda2022neural} toolkit and use the QNN-SDK\footnote{\url{https://developer.qualcomm.com/software/qualcomm-ai-stack}} from Qualcomm\textsuperscript{\faRegistered}
AI Stack.\footnote{Snapdragon and Qualcomm branded products are products of Qualcomm Technologies, Inc. and/or its subsidiaries.}
Table \ref{tab:ondevice} summarizes our evaluation on this target S24 device. Other than the RAFT-S and its variant that run out of memory, an expected behavior due to the all-pair cost volume space consumption for RAFT \cite{teed2020raft} architecture against limited on-target memory, the result shows that our proposed SciFlow method incurs minimal additional overhead in latency and power. 
We observe slightly reduced latency  MobileFlow+SCI+RFL compared to baseline. Though it might seem counter-intuitive, the compiler optimization may be the reason for such observation, an indication for minimal SciFlow latency overhead.


\begin{table}[t]
\begin{center}
\caption{\textbf{On-device evaluation for SciFlow variants over baselines.} We report on-device performance of the baselines, RAFT-S and MobileFlow, and their variants. For fair comparisons, we ensure same on-device execution power mode and apply same number (6) of iterations to meet the real-time requirement for model variants below. "NA/OOM" indicates an out-of-memory error for the model (along with its needed memory for cost volume and activations) during inference. Subsection \ref{subsect:ondevice} has more details.}
\vspace{-8pt}
\label{tab:ondevice}
\adjustbox{max width=0.48\textwidth}
{
\begin{tabular}{|l||c|c|c|}
\hline
Model Architecture & \#Params & Latency (ms) & Power (mW) \\
\hline
\hline
RAFT-S & 1.0M & NA/OOM & NA/OOM \\
RAFT-S+SCI+RFL & 1.0M & NA/OOM & NA/OOM \\
MobileFlow\textsuperscript{\ref{mobileflow}} & 1.5M & 29.02 (+0.00\%) & 392 (+0.00\%)\\
MobileFlow\textsuperscript{\ref{mobileflow}}+SCI+RFL & 1.5M & 28.88 (-0.48\%) & 393 (+0.26\%)\\

\hline
\end{tabular}
}
\vspace{-6mm}
\end{center}
\end{table}




%% file: sec/6_conclusion.tex
\section{Conclusion}
\label{sec:conclusion}
In this paper, we introduce two effective techniques for optical flow estimation. Specifically, we propose Self-Cleaning Iterations (SCI) to help resolve estimation ambiguities, mitigating the issue of error propagation during iterative refinement. Additionally, We propose Regression Focal Loss (RFL) to guide the model to focus on regions of high residual regression errors during training. Our experiments show that SciFlow, the combination of SCI and RFL, significantly improves accuracy of lightweight baseline models at negligible additional overhead for real-time on-device optical flow estimation. We believe our methods may benefit a wider range of model architectures and may be potentially extended to more vision use cases and tasks.